\documentclass{article}
\DeclareUnicodeCharacter{202F}{\,}
\PassOptionsToPackage{numbers}{natbib}

\usepackage[preprint]{neurips_2025}
\usepackage[utf8]{inputenc} 
\usepackage[T1]{fontenc}    
\usepackage{hyperref}       
\usepackage{url}            
\usepackage{booktabs}       
\usepackage{amsfonts}       
\usepackage{nicefrac}       
\usepackage{tikz}
\usepackage{microtype}      
\usepackage{xcolor}         
\usepackage[utf8]{inputenc}
\usepackage{amsmath, amssymb}
\usepackage{graphicx}
\usepackage{hyperref}
\usepackage{pifont}
\usepackage{tikz}
\usepackage{booktabs}
\usepackage{enumitem}
\usepackage{fancyhdr}
\usepackage{hyperref}
\usepackage{titlesec}
\usepackage{microtype}
\usepackage{subcaption}
\usepackage{graphicx}
\usepackage{enumitem}
\usepackage{fancyhdr}
\usepackage{titlesec}
\usepackage{subcaption}
\usepackage{amsmath, amssymb}
\usepackage[most]{tcolorbox} 
\usepackage{wrapfig}
\usepackage{parskip} 
\usepackage[utf8]{inputenc}
\usepackage{graphicx}
\usepackage{booktabs}
\usepackage{multirow}
\usepackage{caption}
\usepackage[numbers]{natbib}
\newtheorem{assumption}{Assumption}

\title{Activation Steering for Chain-of-Thought Compression}

\author{%
  Seyedarmin Azizi\thanks{Corresponding author} \\
  University of Southern California\\
  \texttt{seyedarm@usc.edu} \\
   \And
   Erfan Baghaei Potraghloo \\
   University of Southern California \\
  \texttt{baghaeip@usc.edu} \\
   \AND
   Massoud Pedram \\
   University of Southern California\\
   \texttt{pedram@usc.edu} \\
}

\begin{document}

\maketitle

\begin{abstract}
Large language models (LLMs) excel at complex reasoning when they include intermediate steps, known as \emph{chains of thought} (CoTs). However, these rationales are often overly verbose, even for simple problems, leading to wasted context, increased latency, and higher energy consumption. We observe that verbose, English-heavy CoTs and concise, math-centric CoTs occupy distinct regions in the model’s residual-stream activation space. By extracting and injecting a \emph{steering vector} to transition between these modes, we can reliably shift generation toward more concise reasoning, effectively compressing CoTs without retraining.
We formalize this approach as \textbf{Activation‑Steered Compression (ASC)}, an inference‑time technique that shortens reasoning traces by directly modifying hidden representations. In addition, we provide a theoretical analysis of the impact of ASC on the output distribution, derived from a closed-form KL-divergence-bounded constraint to regulate steering strength. Using only 50 paired verbose and concise examples, ASC achieves up to \textbf{67.43\%} reduction in CoT length on MATH500 and GSM8K datasets, while maintaining accuracy across 7B, 8B, and 32B parameter models. As a training-free method, ASC introduces negligible runtime overhead and, on MATH500, delivers an average \textbf{2.73×} speedup in end-to-end reasoning wall-clock time on an 8B model. This makes ASC a practical and efficient tool for streamlining the deployment of reasoning-capable LLMs in latency- or cost-sensitive settings. The code is available at \href{https://github.com/ArminAzizi98/ASC}{https://github.com/ArminAzizi98/ASC}.
\end{abstract}

\section{Introduction}
Explicit reasoning traces, commonly known as chains of thought (CoTs), significantly enhance the performance of LLMs on multi-step tasks such as mathematical problem solving, logical inference, and program synthesis~\citep{wei2022cot, cobbe2021gsm8k, wang2023selfconsistency}. However, this advantage often comes with the drawback of generating unnecessarily lengthy and verbose rationales~\citep{chen2025think23overthinkingo1like, xu2025chaindraftthinkingfaster}. This verbosity not only increases computational costs by producing more tokens and consuming additional energy, but also risks impairing performance through \emph{overthinking} - where the model introduces redundant steps, multiple self-verifications, and variations~\citep{chen2025think23overthinkingo1like}. This can lead to performance degradation~\citep{wang2025thoughts}. Furthermore, lengthy CoTs pose challenges for deployment in latency-sensitive or resource-constrained environments~\citep{cheng2024compressed}. 

\begin{figure}[ht]
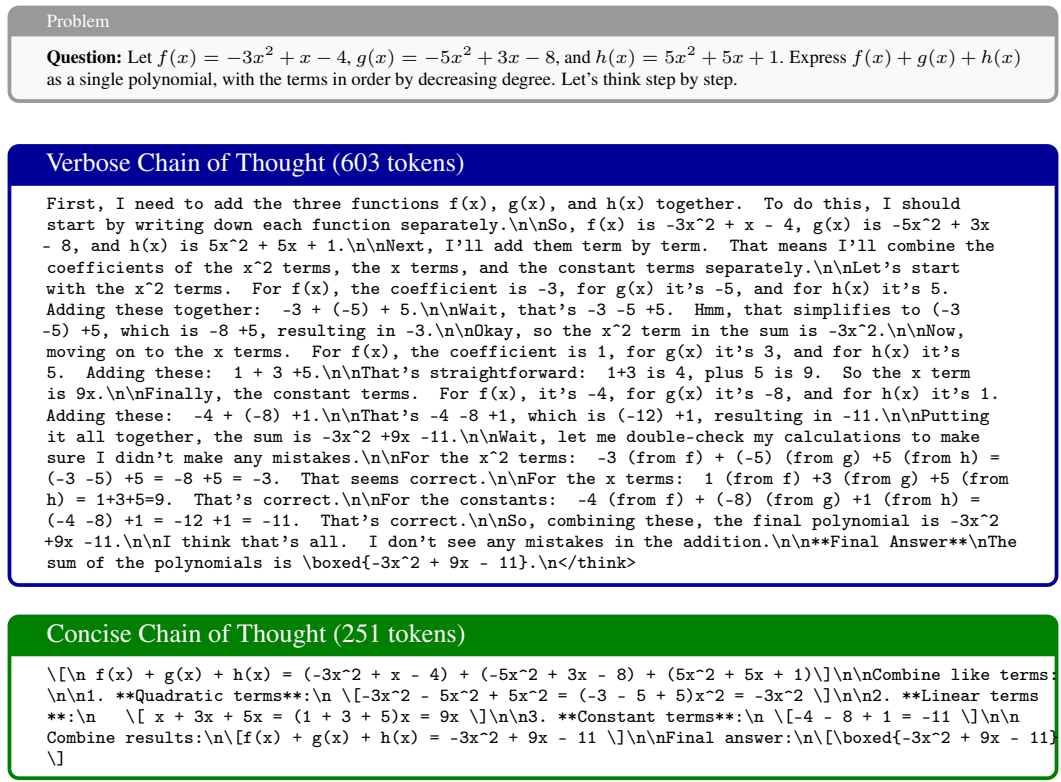

\centering
\scriptsize

\caption{A representative pair of verbose vs. concise CoTs used for generating the steering vector.} 
\label{example_fig}             

\begin{tcolorbox}[
  title=Problem,
  colback=gray!5,
  colframe=black!40,
  width=\textwidth,
  boxsep=2pt, top=2pt, bottom=2pt
]
\textbf{Question:} Let \( f(x) = -3x^2 + x - 4 \), \( g(x) = -5x^2 + 3x - 8 \), and \( h(x) = 5x^2 + 5x + 1 \). Express \( f(x) + g(x) + h(x) \) as a single polynomial, with the terms in order by decreasing degree. Let's think step by step.
\end{tcolorbox}
\end{figure}
\begin{figure}[t]
    \centering
    \includegraphics[width=0.8\linewidth]{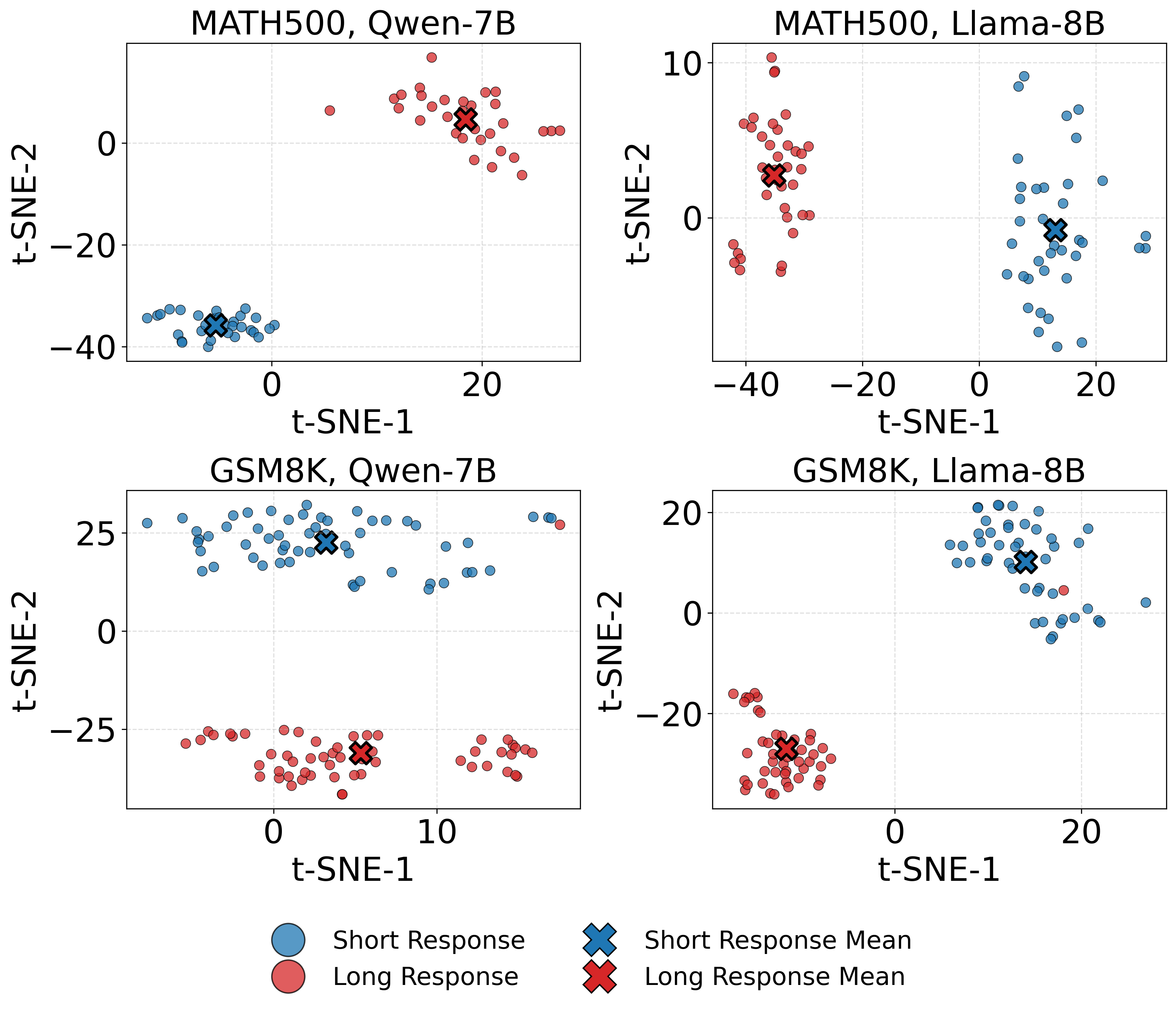}
    \caption{
    t-SNE visualization of residual stream representations for long (verbose) and short (concise) CoT responses across two datasets and two models.
    }
    \label{fig:compact_2x2}
\end{figure}

In this paper, we ask: \emph{Can we compress chains of thought \textbf{without retraining}, by manipulating the model's hidden representations at inference time?} We answer the question affirmatively with \textbf{Activation-Steered Compression (ASC)}. Our key observation is that internal representations of verbose, natural-language CoTs and their concise, math-centric counterparts occupy distinct regions in the model activation space. To evaluate this hypothesis, we sample questions from the \textbf{MATH500}~\citep{hendrycks2021math} and \textbf{GSM8K}~\citep{cobbe2021gsm8k} benchmarks and use two open-weight reasoning models: DeepSeek-Distill-Qwen-7B and DeepSeek-Distill-LLaMA-8B. For each sample, we generate two variants of the CoT: (1) a verbose reasoning chain produced by the model itself under standard prompting and (2) a concise reasoning produced by GPT-4o prompted to minimize natural language verbosity and maximize math-centric reasoning. We feed each input independently into the model and extract residual stream activations, that is, the outputs of the transformer block, in a predetermined layer (e.g., layer 21 in our experiments). A two-dimensional t-SNE projection~\citep{van2008visualizing} of these activations, shown in Figure~\ref{fig:compact_2x2}, reveals a clear separation between the two reasoning styles. This separation motivates the construction of a \textit{steering vector}, which is a direction in the activation space that shifts the model's reasoning towards the concise response generation mode. By extracting this vector from a small calibration set and injecting it during generation, we guide the model to focus on essential steps, reducing verbosity while preserving accuracy. An example of such a pair of responses is shown in Figure~\ref{example_fig}.

\paragraph{Why Activation Steering for CoT Compression?}
Existing methods for compressing CoT reasoning can be broadly categorized into three approaches:
\textbf{(i) retraining-based methods} that fine-tune models to produce shorter rationales, using techniques such as knowledge distillation~\citep{zhang2024ccot} or embedding reasoning within compact latent tokens~\citep{li2024latentcot};
\textbf{(ii) prompt-engineering strategies} that employ carefully designed instructions to encourage models to ``reason briefly,'' utilize contrastive demonstrations, or favor symbolic sketches over verbose prose~\citep{chen2023contrastivecot, yao2023tree}; and
\textbf{(iii) heuristic early-exit mechanisms} that halt generation once a confidence or entropy threshold is reached, trading completeness for speed~\citep{yang2025dynamic}.  
Activation steering offers an intriguing and effective middle ground. It is \emph{lightweight}, requires only the addition of a single vector during inference, and directly reshapes hidden representations to enable on-the-fly compression. Because it does not involve updating the model parameters, this method is \emph{deployment-agnostic}, making it equally applicable to both open-source and closed-source checkpoints. Moreover, it is orthogonal and compatible with the three categories of CoT compression described above. Finally, steering aligns well with interpretability: by nudging hidden states toward the sub-manifold of focused and concise reasoning, it exposes a controllable axis linking latent \emph{representations} to computational \emph{efficiency}.

\begin{tcolorbox}[
  title=Verbose Chain of Thought (603 tokens),
  colback=white,
  colframe=blue!60!black,
  width=\textwidth,
  boxsep=2pt, top=2pt, bottom=2pt
]
\scriptsize
\texttt{First, I need to add the three functions f(x), g(x), and h(x) together. To do this, I should start by writing down each function separately.\textbackslash n\textbackslash nSo, f(x) is -3x\^{}2 + x - 4, g(x) is -5x\^{}2 + 3x - 8, and h(x) is 5x\^{}2 + 5x + 1.\textbackslash n\textbackslash nNext, I'll add them term by term. That means I'll combine the coefficients of the x\^{}2 terms, the x terms, and the constant terms separately.\textbackslash n\textbackslash nLet's start with the x\^{}2 terms. For f(x), the coefficient is -3, for g(x) it's -5, and for h(x) it's 5. Adding these together: -3 + (-5) + 5.\textbackslash n\textbackslash nWait, that's -3 -5 +5. Hmm, that simplifies to (-3 -5) +5, which is -8 +5, resulting in -3.\textbackslash n\textbackslash nOkay, so the x\^{}2 term in the sum is -3x\^{}2.\textbackslash n\textbackslash nNow, moving on to the x terms. For f(x), the coefficient is 1, for g(x) it's 3, and for h(x) it's 5. Adding these: 1 + 3 +5.\textbackslash n\textbackslash nThat's straightforward: 1+3 is 4, plus 5 is 9. So the x term is 9x.\textbackslash n\textbackslash nFinally, the constant terms. For f(x), it's -4, for g(x) it's -8, and for h(x) it's 1. Adding these: -4 + (-8) +1.\textbackslash n\textbackslash nThat's -4 -8 +1, which is (-12) +1, resulting in -11.\textbackslash n\textbackslash nPutting it all together, the sum is -3x\^{}2 +9x -11.\textbackslash n\textbackslash nWait, let me double-check my calculations to make sure I didn't make any mistakes.\textbackslash n\textbackslash nFor the x\^{}2 terms: -3 (from f) + (-5) (from g) +5 (from h) = (-3 -5) +5 = -8 +5 = -3. That seems correct.\textbackslash n\textbackslash nFor the x terms: 1 (from f) +3 (from g) +5 (from h) = 1+3+5=9. That's correct.\textbackslash n\textbackslash nFor the constants: -4 (from f) + (-8) (from g) +1 (from h) = (-4 -8) +1 = -12 +1 = -11. That's correct.\textbackslash n\textbackslash nSo, combining these, the final polynomial is -3x\^{}2 +9x -11.\textbackslash n\textbackslash nI think that's all. I don't see any mistakes in the addition.\textbackslash n\textbackslash n**Final Answer**\textbackslash nThe sum of the polynomials is \textbackslash boxed\{-3x\^{}2 + 9x - 11\}.\textbackslash n</think>}
\end{tcolorbox}

\vspace{0.5em}

\begin{tcolorbox}[
  title=Concise Chain of Thought (251 tokens),
  colback=white,
  colframe=green!50!black,
  width=\textwidth,
  boxsep=2pt, top=2pt, bottom=2pt
]
\scriptsize
\begin{verbatim}
\[\n f(x) + g(x) + h(x) = (-3x^2 + x - 4) + (-5x^2 + 3x - 8) + (5x^2 + 5x + 1)\]\n\nCombine like terms:
\n\n1. **Quadratic terms**:\n \[-3x^2 - 5x^2 + 5x^2 = (-3 - 5 + 5)x^2 = -3x^2 \]\n\n2. **Linear terms
**:\n   \[ x + 3x + 5x = (1 + 3 + 5)x = 9x \]\n\n3. **Constant terms**:\n \[-4 - 8 + 1 = -11 \]\n\n
Combine results:\n\[f(x) + g(x) + h(x) = -3x^2 + 9x - 11 \]\n\nFinal answer:\n\[\boxed{-3x^2 + 9x - 11}
\]
\end{verbatim}
\end{tcolorbox}


Steering involves not only the direction of modifications in the activation space but also selecting the appropriate scale of the steering vector. If the scale is too small, the intervention has little effect; If too large, the model’s output distribution can change unpredictably or even collapse. Previous approaches typically choose this scale heuristically, often by grid search or visual inspection. In contrast, we introduce a principled method for calibrating the steering strength by explicitly bounding the KL divergence between the original and steered output distributions. Our analysis provides a closed-form formula that accounts for both the local sensitivity and curvature of the model’s output relative to the intervention, allowing reliable and provably controlled distribution shifts.

\paragraph{Our key contributions are:}
\begin{enumerate}
\item We conceptualize CoT verbosity as a \emph{latent, steerable dimension} of model behavior, reframing rationale compression as a problem of representation-level control rather than output-level post-processing.

\item We propose \textbf{Activation‑Steered Compression (ASC)}, a training-free framework that uses linear activation injections to reliably shorten CoTs. A single steering vector consistently shifts generations from verbose natural language to concise reasoning chains. ASC is orthogonal to existing compression methods (e.g., early-exit or token pruning) and can be composed with them to further enhance efficiency.

\item We provide the first theoretical framework for safe activation steering by deriving a closed-form scaling rule that bounds the KL divergence at the model output. Our formulation accounts for both the local Jacobian and second-order curvature, enabling principled control over the distributional shift.

\item We conducted extensive experiments in various reasoning tasks and model sizes (7B, 8B, and 32B), showing that ASC reduces CoT length by up to \textbf{67.43\%} on MATH500 and GSM8K without accuracy degradation. On MATH500, ASC delivers an average \textbf{2.73$\times$} inference speedup on an 8B reasoning model, representing, to our knowledge, the largest efficiency gains achieved by a purely inference time intervention.
\end{enumerate}



\section{Background}
We place ASC at the intersection of research on CoT prompting, representation engineering, and the computational economics of long‑context decoding.

\textbf{Chain-of-thought (CoT) prompting} improves multi-step reasoning by encouraging language models to articulate intermediate steps, often using signals such as ``Let's think step by step''~\citep{wei2022chain}. Several enhancements have refined this approach: \textbf{self-consistency}\citep{wang2023self} samples multiple rationales and selects the response supported by the majority; \textbf{tree-of-thought}~\citep{yao2023tree} performs look-ahead search across branching reasoning paths; and \textbf{program-of-thought}~\citep{chen2023program} converts natural language reasoning into executable code. Although effective, these methods often increase the output length significantly. A recent study \citep{chen2025think23overthinkingo1like} showed that o1-style reasoning models frequently produce excessively long CoTs — even for simple questions like ``What is 2 + 3?''— due to redundant computations, unnecessary self-verification, and lingering explanations. We term these inefficiencies \emph{\textbf{verbosity}}, which we aim to address directly through inference-time activation-level intervention.

\textbf{Activation Steering and Representation Engineering}
Linear activation editing has emerged as a lightweight alternative to fine-tuning. Activation Addition (ActAdd) demonstrates that adding a direction corresponding to ``<|positive sentiment|>'' can change the tone of the output~\citep{turner2023actadd}. Reference\citep{burns2024representation} formalizes the approach as \emph{representation engineering}, defining vectors as basis elements in a controllable subspace. The applications now span style transfer~\citep{haviv2024style}, factual correction~\citep{meng2023locating}, and gender debiasing~\citep{liang2023manipulating}. However, to our knowledge, no previous work targets \emph{efficiency} metrics such as token count.

\textbf{Efficiency Challenges in Processing Long Sequences}
Standard decoder-only transformers scale the inference computation with sequence length quadratically. Empirical profiling on an A100 NVIDIA GPU shows that halving a sequence from 8k to 4k tokens reduces latency by \textasciitilde40\% and energy by \textasciitilde35\%. Compression, therefore, offers a direct lever for green AI and a cost-effective deployment~\citep{strubell2019energy}.

\section{Related Work}
Previous work tackles the CoT efficiency gap primarily through methods that \emph{require additional training}: knowledge distillation schemes that learn concise rationales~\citep{zhang2024ccot}, latent token approaches that embed reasoning in compact vectors~\citep{li2024latentcot}, token-level pruning with supervised objectives such as \textsc{TokenSkip}~\citep{xia2025tokenskip}, reinforcement-learning-based trajectory shortening exemplified by \textsc{ThinkPrune}~\citep{hou2025thinkprune}, and latent-reasoning optimization frameworks that fine-tune internal deliberation steps~\citep{chen2025latro}. While effective, these techniques incur considerable computational cost or architectural modifications. In contrast, we propose a \emph{training-free, inference-time} method that compresses CoTs by directly manipulating hidden representations, retaining the accuracy benefits of reasoning traces without the overhead of retraining.

Chain of drafts (CoD)~\citep{xu2025chaindraftthinkingfaster} and the approach of \citep{stolfo2024improving} reduce verbosity by embedding explicit length constraints in the prompt.  CoD instructs the model to “think step by step” but keep the each draft to at most five words, whereas \citep{stolfo2024improving} limits the final answer to a user-specified number of sentences to create inference-time interventions. Although such heuristics can shorten outputs, they assume that the model will faithfully obey length directives, a behavior that recent studies show is unreliable for reasoning-oriented LLMs~\citep{fu2025mathif}.
The closest work to ours is SEAL~\citep{chen2025seal}, which constructs its steering vector by manually labeling the thought segments as \emph{execution}, \emph{reflection}, or \emph{transition}, and then damping the latter two segment types. In contrast, (i) we learn a single \emph{verbosity axis} from paired \textsc{verbose}–vs.–\textsc{concise} CoTs without any manual labels, (ii) rely solely on off-the-shelf prompts to generate training pairs, and (iii) obtain a domain-agnostic vector that generalizes across reasoning tasks. Therefore, our method provides a taxonomy-free, training-free complement to SEAL’s category-based calibration.


\begin{figure}[!t]
    \centering
    \includegraphics[width=0.90\linewidth]{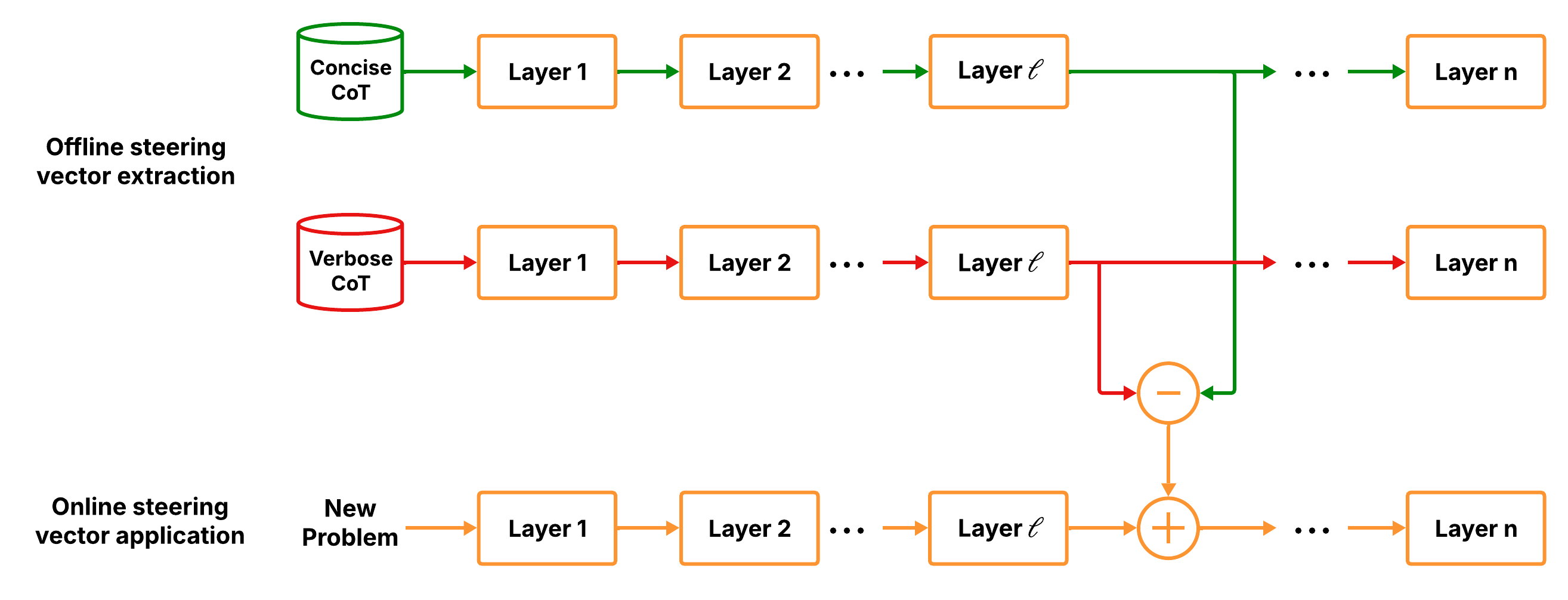}
    \caption{Steering vector extraction and application using pairs of concise and verbose CoTs.}
    \label{fig:method}
\end{figure}
\section{Activation‑Steered Compression}
\label{sec:asc}




Motivated by the goal of improving CoT efficiency through manipulation of the model's activation space, we introduce \textbf{Activation-Steered Compression (ASC)}—a method that shifts the model's hidden representations toward the subspace associated with concise, math-centric chains of thought. The method is summarized in Figure \ref{fig:method}. First, we randomly sample 50 calibration samples from target dataset (in our case we have focused on \textbf{MATH500} \citep{hendrycks2021math} and from \textbf{GSM8K} \citep{cobbe2021gsm8k}). For each question $q_i$ in the calibration set, we obtain:
\begin{itemize}
    \item \textbf{Verbose CoT} $l_i$ – generated by the target model with standard CoT prompting \citep{wei2022cot}.
    \item \textbf{Concise CoT} $s_i$ – produced by GPT‑4o instructed to use concise math-centric reasoning with minimal English.
\end{itemize}

We denote the output of the transformer block in layer $\ell$ as \emph{residual stream} of layer $\ell$, and use $h^\ell$ to refer to it. Formally, $h^\ell$ is a matrix of shape $T \times d$, where $T$ is the number of tokens in the input sequence and $d$ is the hidden dimension of the model. With a slight abuse of notation, we write $h^\ell(s)$ to denote the residual stream in layer $\ell$ when a string $s$ is fed into the model.

Following this notation, we feed the concatenated input \texttt{[question + CoT]} into the target model and extract the residual-stream activations corresponding to the final token in the input sequence. Specifically, we obtain $h^\ell(q_i \oplus l_i)[-1]$\footnote{$\oplus$ is the string concatenation operator.} and $h^\ell(q_i \oplus s_i)[-1]$ at a selected layer $\ell$, corresponding to the verbose and concise CoTs, respectively. The steering vector is then computed as the average difference between these final token activations, that is, the activations associated with the token $\texttt{EOS}$, in all examples $N$.

\[
v^\ell = \frac{1}{N} \sum_{i=1}^N \left(h^\ell(q_i \oplus s_i)[-1] - h^\ell(q_i \oplus l_i)[-1]\right).
\]

\(v^\ell\) is the desired steering direction for shifting the long and verbose CoT toward a more compact CoT. At inference time, given a new question and the \(i\)-th generated token \(x_i\), we modify the residual stream by injecting the steering vector \(v^\ell\) into layer \(\ell\) during each decoding step, until an end-of-sequence delimiter is emitted. Specifically, for each decoding step \(i\), we apply:

\[
    h^\ell(x_i) \leftarrow h^\ell(x_i) + \gamma v^\ell \quad \forall\, i \in [1,\, \texttt{decoding\_steps}]
\]

Here, $\gamma$ is a hyperparameter that controls the injection strength of the steering vector. If $\gamma$ is too large, it can significantly distort the residual stream distribution, leading to degenerate or incoherent output. In contrast, if $\gamma$ is too small, the steering effect becomes negligible. In the remainder of this section, we analyze the effect of the scaling parameter \(\gamma\) on the model output distribution from a theoretical perspective. We derive a safe upper bound on \(\gamma\) that guarantees that the output divergence remains within a user-specified threshold.

For simplicity, we now drop the layer index $\ell$ throughout the analysis. We steer hidden activations by adding a direction~$\mathbf v$ at layer~$\ell$, and choose the scale~$\gamma$ so that the resulting output distribution remains close to the unsteered model. Formally, letting $\mathbf z$ and $\tilde{\mathbf z}$ denote the pre-softmax logits before and after steering, we constrain the forward KL divergence:
\[
\mathrm{KL}\!\bigl(\operatorname{softmax}(\mathbf z)
\;\Vert\;
\operatorname{softmax}(\tilde{\mathbf z})\bigr)
\;\le\;\varepsilon,
\]
where $\varepsilon$ is a user-specified divergence budget (we use $\varepsilon=10^{-3}$ in practice).

The full derivation, deferred to Appendix~\ref{sec:proof_kl_corrected}, decomposes the logit shift into a linear component $\gamma\mathbf W\mathbf v$ and a curvature-dependent remainder, where $\mathbf W$ is the Jacobian \(\mathbf{J(.)}\) of the logit map with respect to the activations of layer \(\ell\). Under a mild smoothness condition with constant $L$ as the upper bound of \emph{directional curvature}, we derive a provable upper bound of KL that is quadratic, cubic, and quartic in $\gamma$. Specifically, defining
\[
a := \bigl\|\mathbf W\mathbf v\bigr\|_2,
\qquad
L := \sup_{t\in[0,\gamma]}\!
      \frac{\|\mathbf J(\mathbf h+t\mathbf v)-\mathbf J(\mathbf h)\|_2}{t},
\]
we obtain a closed-form scale $\gamma_{\max}$ that ensures $\mathrm{KL}\le\varepsilon$. The expression includes a curvature-aware safety factor:
\[
\gamma_{\max} =
\max\!\left\{0,\left(1-\tfrac{L\gamma_{\mathrm{raw}}}{4a}\right)\gamma_{\mathrm{raw}}\right\},
\]
where $\gamma_{\mathrm{raw}} = (a/L)\cdot x$ and \(x\) is determined by solving the dimensionless cubic equation \(x^{3}+x^{2}-4\varepsilon L^{2}/a^{4}=0\). All constants are explicit; no additional hyperparameters are introduced.

In practice, we estimate the two scale parameters \(\alpha\) and \(L\) on the small calibration set (50 hidden states).  
For each hidden state, we evaluate one Jacobian--vector product with the chosen steering direction and record its Euclidean norm; the median of these norms is taken as our estimate of \(a\).  
To obtain \(L\), we compute a single Hessian vector product along the same direction at each calibration point, collect the resulting norms, take their \(95^{\text{th}}\) percentile. Both JVP and HVP operations are one‐line calls in modern autodiff frameworks, so the entire procedure runs in a few seconds even on large-scale models.

\medskip


All proofs, derivations, and bounds appear in Appendix~\ref{sec:proof_kl_corrected}. We adopt this calibrated $\gamma_{\max}$ in all experiments to control distributional shift while preserving the intended compression effect of each steering vector.

\section{Experiments}

This section presents our experimental results demonstrating that ASC effectively reduces the length of CoT reasoning while maintaining or improving task performance. We begin by describing our experimental setup in section~\ref{sec:setupp}, followed by the main results in section~\ref{sec:mainn}.

\subsection{Experimental Setup}
\label{sec:setupp}
\textbf{Models, Datasets, and Baselines.} We evaluate ASC on several recent open-source reasoning models: DeepSeek-R1-Distill-LLaMA-8B \citep{deepseekai2025deepseekr1incentivizingreasoningcapability}, DeepSeek-R1-Distill-Qwen-7B \citep{deepseekai2025deepseekr1distillqwen7b}, and QwQ-32B \citep{qwen2025qwq32b}. The evaluation is performed on multiple reasoning benchmarks, including MATH-500 \citep{hendrycks2021measuring} and GSM8K \citep{cobbe2021gsm8k}. As baselines, we compare ASC against vanilla CoT prompting (no steering), CoD \citep{xu2025chaindraftthinkingfaster}, DEER \citep{yang2025dynamic}, TCC \citep{muennighoff2025s1}, and SEAL \citep{chen2025seal}, a recent method for compressed reasoning that uses steering vectors.

\textbf{Implementation Details.}  
For all experiments, we use the decoding hyperparameters \(\texttt{temperature} = 0.7\), \(\texttt{top\_p} = 0.9\), and \(\texttt{repetition\_penalty} = 1.1\); all other settings follow the default configurations of the respective models. The evaluation datasets are accessed through the Hugging Face \texttt{datasets} library. Experiments are conducted on NVIDIA A6000 GPUs, using \texttt{PyTorch} version \texttt{2.5.1+cu124} and the \texttt{transformers} library version \texttt{4.50.1}. The hyperparameters related to steering, namely the steering strength \(\gamma\) and the layer index used to extract and apply the steering vector, are detailed in the appendix~\ref{append:hypers}.

\subsection{Main Results}
\label{sec:mainn}
Table~\ref{tab:main} presents the performance of ASC compared to baseline CoT compression techniques. On the DeepSeek-R1-Distill-LLaMA-8B model, ASC reduces CoT length by up to \textbf{61.2\%} without any loss in accuracy, outperforming prior methods in compression effectiveness. On the same model and the GSM8K dataset, ASC achieves a compression rate of \textbf{67.43\%}, while also slightly improving answer accuracy by \textbf{0.2\%}, matching or exceeding the performance of the vanilla CoT baseline. On MATH500, ASC achieves a \textbf{33.8\%} reduction in CoT length, again outperforming all baselines while maintaining equivalent accuracy.

On the larger QwQ-32B model, ASC compresses CoTs by \textbf{50.7\%} and \textbf{45.7\%} on MATH500 and GSM8K, respectively. Notably, on MATH500, it also yields a \textbf{0.4\%} accuracy improvement over the vanilla CoT. Upon inspection, we find that the high token count in some model responses arises primarily from either examples exceeding their token budget or exhibiting excessive branching and thought switching during generation. This aligns with the observations of~\citep{wang2025thoughts}, who show that LLMs similar to o1 tend to generate longer responses when frequently switching between reasoning paths without deeply pursuing any one. This behavior, termed \textit{ under-thinking}, often manifests itself as verbose outputs filled with abandoned or partially developed reasoning trajectories. Among the models evaluated, QwQ-32B appears particularly susceptible to this issue. On the challenging MATH500 benchmark, ASC mitigates this behavior by promoting concise, linear reasoning and earlier halting, thereby suppressing extraneous chains of thought. Qualitative examples that illustrate this suppression are provided in Appendix~\ref{append:qualitative}, where ASC responses exhibit significantly fewer thought changes than their vanilla CoT counterparts. In summary, across all models and datasets, ASC consistently achieves the \textbf{highest CoT compression} while preserving the final answer accuracy.

\begin{wrapfigure}{r}{0.51\textwidth}
  \begin{center}
    \includegraphics[width=0.5\textwidth]{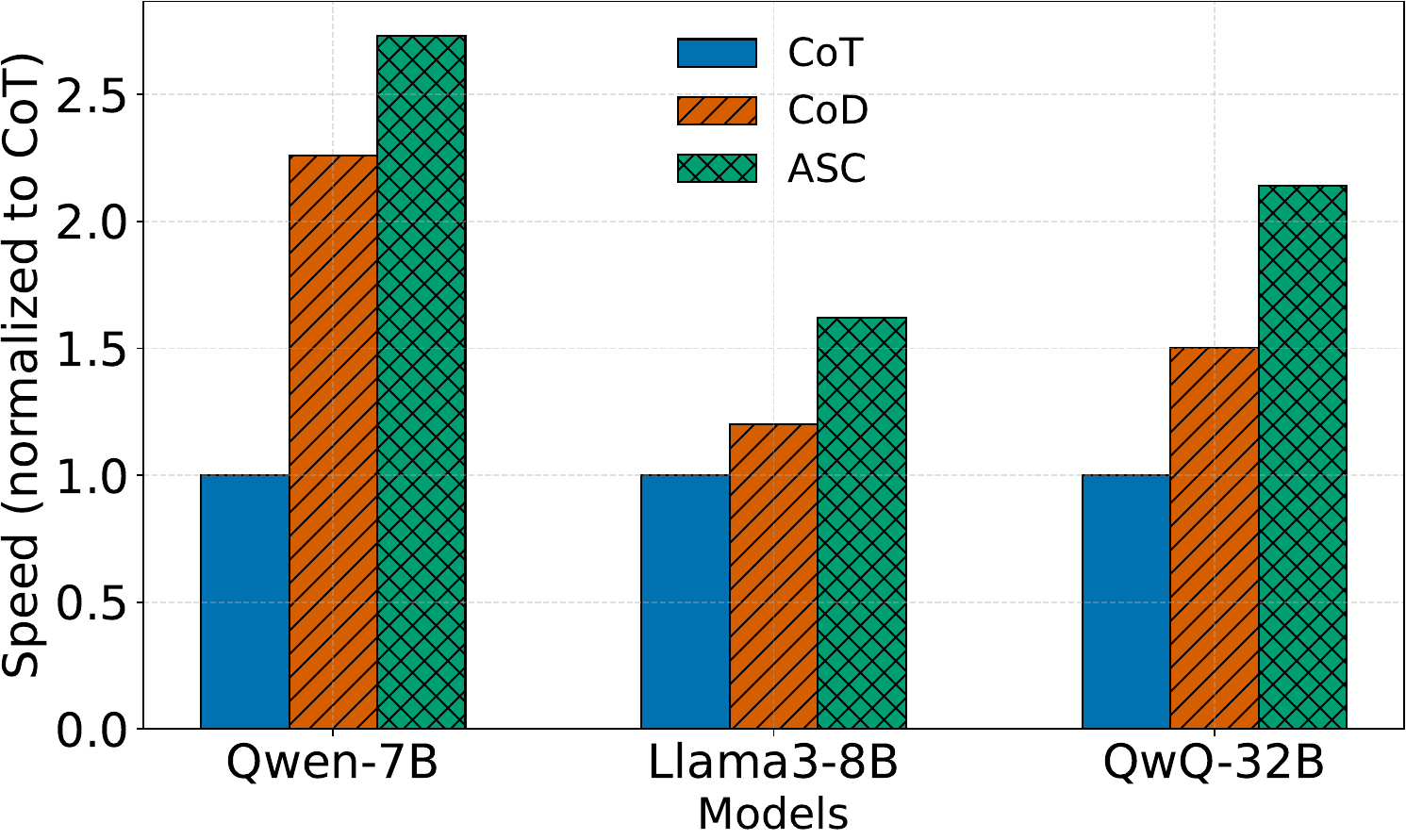}
  \end{center}
  \caption{Speed comparison of CoT, CoD, and ASC on MATH500 dataset.}
  \label{fig:speed}
\end{wrapfigure}

Since one of the primary goals of CoT compression is to reduce end-to-end response latency, we measure the average generation time for three models—DeepSeek-R1-Distill-LLaMA-8B, DeepSeek-R1-Distill-Qwen-7B, and QwQ-32B—on the MATH500 dataset. Latency is measured on an NVIDIA A6000 GPU. We then compute and report the inverse latency (i.e., generation speed) for three decoding strategies: standard CoT, Chain-of-Drafts (CoD), and our proposed ASC, as shown in Figure~\ref{fig:speed}. The results indicate that ASC improves the generation speed of CoT-based reasoning by up to \(2.73\times\), with no loss in answer accuracy.

\begin{table}[ht]
\centering
\caption{Performance comparison of CoT, TCC, DEER, CoD, SEAL and ASC on reasoning tasks.}
\resizebox{\textwidth}{!}{
\begin{tabular}{cc|cc|cc}
\toprule
\multirow{2}{*}{\textbf{Model}} & \multirow{2}{*}{\textbf{Method}} 
& \multicolumn{2}{c|}{\textbf{MATH500}} 
& \multicolumn{2}{c}{\textbf{GSM8k}} \\
\cmidrule{3-6}
& & Acc. (\%) $\uparrow$ & Tokens $\downarrow$ 
  & Acc. (\%) $\uparrow$ & Tokens $\downarrow$ \\
\midrule
\multirow{6}{*}{Deepseek-R1-Distill-Qwen-7B} 
& CoT  & 88.8 & 3984 & 88.6 & 1080 \\
& TCC  & 89.2 & 3864  & 88.0 & 892 \\
& DEER & \textbf{89.8} & 2143 & \textbf{90.6} & 917 \\
& SEAL & 89.4 & 2661 & 88.4 & 811 \\
& CoD  & 88.2 & 1852 & 87.9 & 550 \\
& ASC  & 89.0 & \textbf{1543} & 88.6 & \textbf{536} \\
\midrule
\multirow{4}{*}{Deepseek-R1-Distill-LLaMA-8B} 
& CoT  & 89.2 & 3554 & 89.1 & 2610 \\
& DEER & 89.2 & 2830  & 89.3 & 2124 \\
& CoD  & 88.8 & 3028 & 89.1 & 914 \\
& ASC  & \textbf{89.2} & \textbf{2353}  & \textbf{89.3} & \textbf{850} \\
\midrule
\multirow{5}{*}{QwQ-32B} 
& CoT  & 93.8 & 4508 & \textbf{96.5} & 1530 \\
& TCC  & 94.4 & 4315  & 95.8 & 1348 \\
& DEER & \textbf{94.6} & 3316  & 96.3 & 977 \\
& CoD  & 93.8 & 3400 & 96.2 & 1116 \\
& ASC  & 94.2 & \textbf{2222}  & 96.4 & \textbf{830} \\
\bottomrule
\end{tabular}
} 
\label{tab:main}
\end{table}

\section{Discussion and Ablations}
\textbf{Cross-Task Generalization.} \\ To investigate whether CoT verbosity is consistently reflected in the model's representation space, we examine the alignment of ASC steering vectors extracted from different reasoning tasks. Specifically, we analyze whether steering vectors derived from one dataset generalize to another. We conduct this study using the DeepSeek-R1-Distill-Qwen-7B model and two benchmarks: GSM8K and MATH500. Following the ASC methodology, we independently compute steering vectors for each dataset using 50 paired examples. We then assess the cosine similarity between the two vectors to quantify their alignment. In addition, we evaluate cross-task generalization by applying each dataset's steering vector to compress CoTs in the other dataset, measuring both length reduction and accuracy retention. The results are presented in
Table~\ref{tab:crossdomain}. 
First, the cosine similarity between the two steering vectors is \(\textbf{0.92}\), indicating strong alignment in the vectors from verbose to concise CoTs in MATH500 and GSM8K. Second, the performance of cross-dataset steering matches closely that of in-dataset vectors. Although there is a slight drop in accuracy and a slight increase in token count, ASC with cross-dataset steering still outperforms the vanilla CoT baseline (Table~\ref{tab:main}). These findings suggest that verbosity reduction occupies a \textit{largely shared} latent direction across reasoning tasks, supporting our initial hypothesis that CoT efficiency can generally be attributed to the latent representations of the model.
\begin{table}[h!]
\centering
\begin{tabular}{lcccc}
\toprule
\textbf{Dataset} & \textbf{Steering Vector Source} & \textbf{Accuracy (\%)} & \textbf{CoT Tokens} \\
\midrule
\multirow{2}{*}{MATH500} 
& MATH500 (in-dataset) & 89.0 & 1543 \\
& GSM8K (cross-dataset) & 88.8 & 1631 \\
\midrule
\multirow{2}{*}{GSM8K} 
& GSM8K (in-dataset) & 88.6 & 536 \\
& MATH500 (cross-dataset) & 88.4 & 611 \\
\bottomrule
\end{tabular}
\vspace{1em}
\caption{Performance of ASC on MATH500 and GSM8K using dataset-specific vs.\ cross-dataset steering vectors. The model used is DeepSeek-Distill-Qwen-7B.}
\label{tab:crossdomain}
\end{table}

\begin{wrapfigure}{r}{0.51\textwidth}
  \begin{center}
    \includegraphics[width=0.5\textwidth]{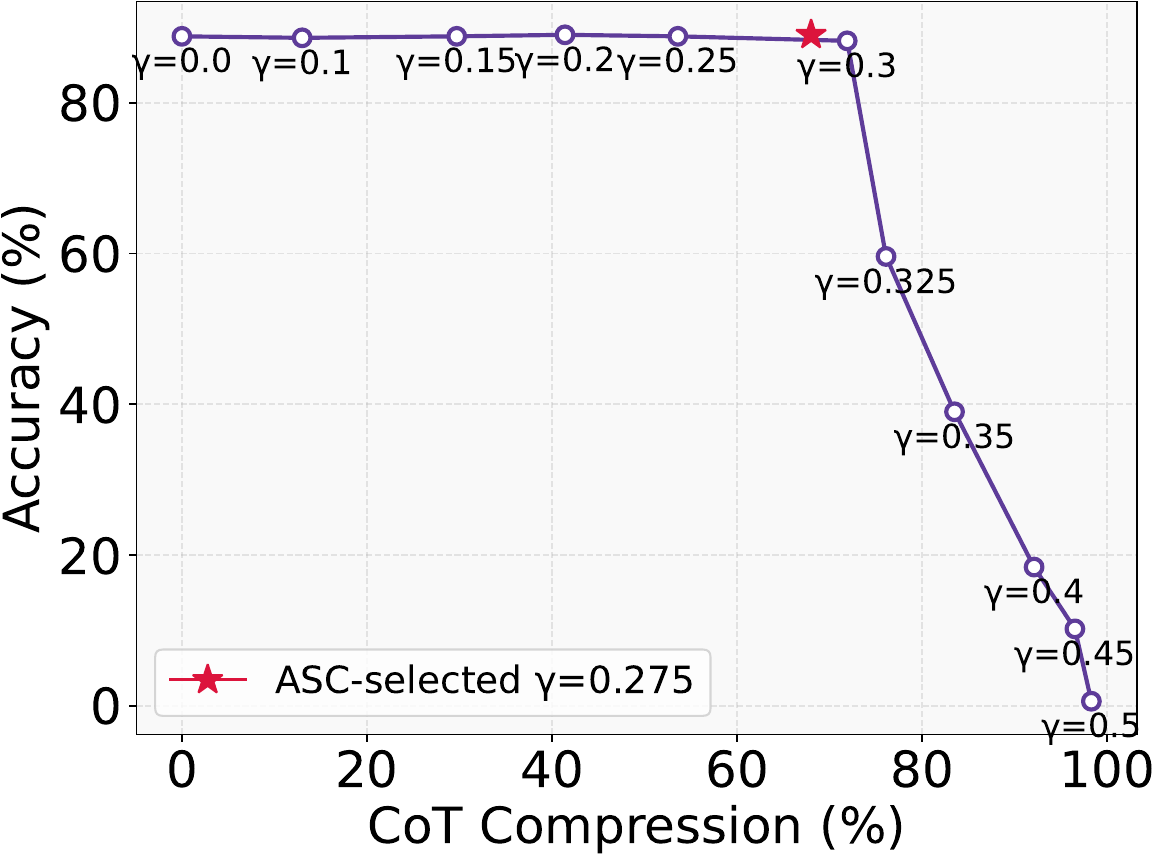}
  \end{center}
  \caption{Effect of steering strength \(\gamma\) on CoT compression and answer accuracy for the DeepSeek-R1-Distill-Qwen-7B model on the MATH500 dataset.
}
  \label{fig:gamma}
\end{wrapfigure}

\textbf{Effect of Steering Strength \(\gamma\).}  
\\The steering strength \(\gamma\) is a critical hyperparameter in ASC, as it directly influences both the degree of CoT compression and the quality of the generated output. To analyze its effect, we use the DeepSeek-R1-Distill-Qwen-7B model on the MATH500 dataset and perform a sweep over a range of \(\gamma\) values. The sweep begins at \(\gamma = 0\) (i.e., no steering) and gradually increases until the steering induces noticeable compression along with a significant drop in answer accuracy \(\gamma = 0.5\). The results are shown in Figure~\ref{fig:gamma}, highlighting the trade-off between CoT compression and answer accuracy as the steering strength \(\gamma\) increases. For small values of \(\gamma\), increasing the strength yields substantial reductions in CoT length with minimal impact on accuracy. However, beyond a certain point, further increases in \(\gamma\) lead to significant accuracy degradation despite continued compression. Notably, the value of \(\gamma\) selected by ASC—computed via the KL-divergence–constrained scaling described in Section~\ref{sec:asc}—closely aligns with the empirical breakpoint where performance begins to degrade. This supports the theoretical grounding of our method for setting steering strength.

\section{Conclusion}
We introduce Activation-Steered Compression (ASC), a training-free method for reducing the verbosity of Chain-of-Thought (CoT) reasoning in large language models by manipulating internal representations at inference time. By leveraging steering vectors derived from paired verbose and concise rationales, ASC effectively compresses CoTs without sacrificing accuracy. We further contribute a closed-form, KL-constrained scaling framework for principled control of steering strength, and provide empirical evidence that verbosity lies along a shared latent direction across tasks. ASC complements existing CoT compression techniques and requires no retraining, and overall advances the efficiency and practicality of LLM-based reasoning by showing that conciseness is not only desirable but also steerable via the internal geometry of the model.

\bibliography{neurips_2025}
\bibliographystyle{neurips_2025}


\appendix
\section{KL-Constrained Scaling of Steering Vectors}
\subsection{Bounding the Distributional Shift of Additive Steering }\label{sec:proof_kl_corrected}

We study the output--distribution shift incurred when an \emph{additive steering} update is applied to the hidden state at layer~$\ell$ of a language model. For an activation vector $\mathbf h\in\mathbb R^d$ we form
\[
 \tilde{\mathbf h}:=\mathbf h+\gamma\mathbf v,\qquad\|\mathbf v\|_2=1,
\]
to analyze how large the Kullback--Leibler (KL) divergence between the pre- and post-steering output distributions can become.

Throughout, let $\mathcal F_{l\to\mathrm{logit}}\!:\mathbb R^d\to\mathbb R^m$ denote the sub-network that maps layer-$\ell$ activations to the pre-softmax \emph{logits}. All vector norms $\|\cdot\|_2$ and operator-$2$ norms are Euclidean; they coincide when the argument is a vector.

\paragraph{Notation for higher-order derivatives.} The Jacobian of $\mathcal F_{l\to\mathrm{logit}}$ at $\mathbf h$ is the matrix
\[
 \mathbf J(\mathbf h):=\nabla_{\mathbf h}\mathcal F_{l\to\mathrm{logit}}(\mathbf h)\in\mathbb R^{m\times d},
\]
whose $j$-th row is $(\nabla_{\mathbf h}F_j(\mathbf h))^{\!\top}$. The Hessian of a scalar component is the usual matrix of second partials. For a unit vector $\mathbf a$ we abbreviate \emph{directional} Hessian Evaluation
\[
 \nabla^{2}_{\mathbf h}\mathcal F_{l\to\mathrm{logit}}(\mathbf h)[\mathbf a,\mathbf a] :=\bigl(\nabla^{2}_{\mathbf h}F_1(\mathbf h)[\mathbf a,\mathbf a],\dots,\nabla^{2}_{\mathbf h}F_m(\mathbf h)[\mathbf a,\mathbf a]\bigr)^{\!\top}\in\mathbb R^{m}.
\]

\subsubsection{A smoothness assumption}

\begin{assumption}\label{assu:smooth}
There exists a constant $L>0$ such that for every unit direction $\mathbf v$ and every $t\in[0,\gamma]$
\[
 \bigl\|\mathbf J(\mathbf h+t\mathbf v)-\mathbf J(\mathbf h)\bigr\|_2\le L\,t.
\]
\end{assumption}

\paragraph{Implication.} Assumption~\ref{assu:smooth} is stronger than merely requiring bounded second derivatives. In fact, according to the mean value theorem for vector-valued Lipschitz maps, $\mathbf J$ is differentiable almost everywhere and its derivative (the third-order tensor of second partials) has the operator norm at most~$L$. Contracting this tensor twice with the same unit vector $\mathbf v$ yields
\begin{equation}\label{eq:dir_hess_bound}
\bigl\|\nabla^{2}_{\mathbf h}\mathcal F_{l\to\mathrm{logit}}(\mathbf h+\tau\mathbf v)[\mathbf v,\mathbf v]\bigr\|_2\le L,\qquad\forall\tau\in[0,\gamma],
\end{equation}
because $\|H[\mathbf v,\mathbf v]\|_2\le\|H\|_{\mathrm{op}}\|\mathbf v\|_2^{2}=\|H\|_{\mathrm{op}}$.
Thus Assumption~\ref{assu:smooth} \emph{implies}--though it is not equivalent to--a uniform bound on the directional Hessian.

\subsubsection{Local linearization with a controlled remainder}

Define
\[
 \mathbf z:=\mathcal F_{l\to\mathrm{logit}}(\mathbf h),\qquad
 \mathbf W:=\mathbf J(\mathbf h)\in\mathbb R^{m\times d}.
\]
By the fundamental theorem of calculus and Eq.~\eqref{eq:dir_hess_bound}, the steered logits decompose as
\begin{align}
\tilde{\mathbf z}
&= \mathcal F_{l\to\mathrm{logit}}(\mathbf h+\gamma\mathbf v)\\[3pt]
&= \mathbf z
   +\underbrace{\gamma\mathbf W\mathbf v}_{:=\boldsymbol\delta}
   +\underbrace{\int_{0}^{\gamma}(\gamma-s)
       \nabla^{2}_{\mathbf h}\mathcal F_{l\to\mathrm{logit}}
       (\mathbf h+s\mathbf v)[\mathbf v,\mathbf v]\,ds}_{:=\mathbf r(\gamma)}.
\end{align}
The \emph{linear component} is $\boldsymbol\delta=\gamma\mathbf W\mathbf v$, while the \emph{remainder} obeys
\begin{equation}\label{eq:r_bound}
\|\mathbf r(\gamma)\|_2\le\tfrac12\,L\gamma^{2}.
\end{equation}

\subsubsection{KL divergence as a Bregman divergence}

Let $g(\mathbf x)=\log\sum_{i=1}^{m}e^{x_i}$ and denote
\[
 \mathbf p   = \operatorname{softmax}(\mathbf z),\qquad
 \tilde{\mathbf p} = \operatorname{softmax}(\tilde{\mathbf z}).
\]
For the log-partition function $g$, the Bregman divergence is
\begin{equation}\label{eq:kl_bregman}
D_g(\tilde{\mathbf z},\mathbf z)
  = g(\tilde{\mathbf z}) - g(\mathbf z)
    - \langle\nabla g(\mathbf z),\tilde{\mathbf z}-\mathbf z\rangle
  = \mathrm{KL}(\mathbf p\Vert\tilde{\mathbf p}).
\end{equation}
Thus, the classical \emph{forward} KL direction appears.

\subsubsection{Integral representation and spectral bound}

Using the integral representation of a Bregman divergence for twice-differentiable convex~$g$ we obtain
\begin{equation}\label{eq:kl_integral}
\mathrm{KL}(\mathbf p\Vert\tilde{\mathbf p})
  = \int_{0}^{1}(1-t)
      (\tilde{\mathbf z}-\mathbf z)^{\!\top}
      \nabla^{2} g\bigl(\mathbf z+t(\tilde{\mathbf z}-\mathbf z)\bigr)
      (\tilde{\mathbf z}-\mathbf z)\,dt.
\end{equation}
Because $\nabla^{2} g(\mathbf x)$ equals the Fisher information matrix
$\mathbf F(\mathbf x)=\operatorname{diag}(\mathbf p)-\mathbf p\mathbf p^{\!\top}$,
whose largest eigenvalue never exceeds $\tfrac12$, and the factor $(1-t)$ integrates to $\tfrac12$, we have
\begin{equation}\label{eq:kl_quadratic}
\mathrm{KL}(\mathbf p\Vert\tilde{\mathbf p})
   \le \tfrac14\,\|\tilde{\mathbf z}-\mathbf z\|_2^{2}.
\end{equation}
This constant $1/4$ is tight for our purposes.

\subsubsection{Putting the pieces together}

With $\tilde{\mathbf z}-\mathbf z=\boldsymbol\delta+\mathbf r(\gamma)$ and the triangle inequality,
\begin{align}
\|\tilde{\mathbf z}-\mathbf z\|_2^{2}
  &\le\bigl(\|\boldsymbol\delta\|_2+\|\mathbf r(\gamma)\|_2\bigr)^{2}\\[3pt]
  &\le \|\boldsymbol\delta\|_2^{2}
       + 2\|\boldsymbol\delta\|_2\|\mathbf r(\gamma)\|_2
       + \|\mathbf r(\gamma)\|_2^{2}.
\end{align}
Invoking~\eqref{eq:r_bound} and $\|\boldsymbol\delta\|_2=\gamma a$ with
$a:=\|\mathbf W\mathbf v\|_2$, we derive from~\eqref{eq:kl_quadratic} the
\textbf{corrected steering bound}:
\begin{equation}\label{eq:full_bound}
\boxed{\mathrm{KL}(\mathbf p\Vert\tilde{\mathbf p})
   \le \tfrac14\,\gamma^{2}a^{2}
      +\tfrac14\,L a\,\gamma^{3}
      +\tfrac1{16}\,L^{2}\gamma^{4}}
\end{equation}

\paragraph{Safe \boldmath$\gamma$ budget with a curvature safety factor.}

Fix a target divergence $\varepsilon>0$. Ignoring the last term in~\eqref{eq:full_bound} yields the cubic inequality
\[
 \tfrac14 a^{2}\gamma^{2}+\tfrac14 L a\,\gamma^{3}\le\varepsilon.
\]
Set $x:=(L\gamma)/a$ (dimensionless) and $\beta:=4\varepsilon L^{2}/a^{4}$. The inequality becomes
\[
 x^{3}+x^{2}-\beta\le0,
\]
whose \emph{unique positive} root solves
\[
 x^{3}+x^{2}-\beta=0.
\]
Writing the depressed cubic
\[
 (x+\tfrac13)^{3}-\tfrac13(x+\tfrac13)+\bigl(\tfrac{2}{27}-\beta\bigr)=0
\]
and setting
\[
 p=-\tfrac13,\qquad q=\tfrac{2}{27}-\beta,\qquad
 \Delta=\bigl(\tfrac{q}{2}\bigr)^{2}+\bigl(\tfrac{p}{3}\bigr)^{3},
\] 
the real Cardano root is
\begin{equation}\label{eq:x_root}
x=\sqrt[3]{-\tfrac{q}{2}+\sqrt{\Delta}}+
  \sqrt[3]{-\tfrac{q}{2}-\sqrt{\Delta}}-
  \tfrac13.
\end{equation}
Numerically, this expression is unambiguous if one takes the \emph{real} branch of each cube root. Finally,
\begin{equation}\label{eq:gamma_raw}
\boxed{\gamma_{\mathrm{raw}}=\frac{a}{L}\,x.}
\end{equation}

\paragraph{Degenerate direction $a=0$.}  
If $a=0$ (the steering vector lies in the null-space of $\mathbf W$) the quadratic and cubic terms vanish; retaining the quartic term in~\eqref{eq:full_bound} gives $\tfrac{L^{2}}{16}\gamma^{4}\le\varepsilon$ and hence $\gamma\le(16\varepsilon)^{1/4}/\sqrt{L}$. We therefore set
\[
 \gamma_{\mathrm{raw}} =
   \begin{cases}
     (a/L)\,x & \text{if } a>0,\\[4pt]
     (16\varepsilon)^{1/4}/\sqrt{L} & \text{if } a=0.
   \end{cases}
\]

\paragraph{Curvature safety factor.}
Because the quartic term in~\eqref{eq:full_bound} is strictly positive, $\gamma_{\mathrm{raw}}$ is slightly optimistic when $L\gamma$ is not negligible relative to~$a$. We therefore define the final scale
\begin{equation}\label{eq:gamma_max}
\boxed{\gamma_{\max}=
  \max\!\bigl\{0,\bigl(1-\tfrac{L\gamma_{\mathrm{raw}}}{4a}\bigr)\gamma_{\mathrm{raw}}\bigr\}}
\end{equation}
with the convention $L\gamma_{\mathrm{raw}}/a:=0$ when $a=0$. One checks directly that substituting $\gamma=\gamma_{\max}$ into~\eqref{eq:full_bound} yields $\mathrm{KL}\le\varepsilon$ whenever $x<4$, which covers all practical regimes.

When $L\to0$ the safety factor tends to~$1$ and Eq.~\eqref{eq:gamma_max} continuously reduces to the familiar local-linear scale $\gamma_{\max}=2\sqrt{\varepsilon}/a$.


\section{Qualitative Results}
\label{append:qualitative}
In this section, we present illustrative examples from the MATH500 dataset, comparing standard Chain-of-Thought (CoT) responses with those produced by ASC. Figures~\ref{example_fig_2} and~\ref{example_fig_3} showcase representative outputs from the QwQ-32B reasoning model. As noted in prior work~\citep{wang2025thoughts, chen2025think23overthinkingo1like}, vanilla CoT traces often exhibit excessive verbosity, frequent shifts in reasoning strategy, and redundant self-verification steps. In contrast, ASC responses are more focused, mathematically grounded, and free from distractive interjections such as \textit{``Alternatively,''} \textit{``Wait,''} and other similar expressions. This reflects ASC’s ability to steer the model toward concise, mathematically structured reasoning.

\begin{figure}[ht]
\centering
\caption{Qualitative Example for comparing ASC response against vanilla CoT response.} 
\label{example_fig_2}             

\begin{tcolorbox}[
  title=Problem,
  colback=gray!5,
  colframe=black!40,
  width=\textwidth,
  boxsep=2pt, top=2pt, bottom=2pt
]
\scriptsize
\textbf{Question:}The product of two consecutive positive even integers is 288. What is the greater of the two integers?
\end{tcolorbox}
\end{figure}

\begin{tcolorbox}[
  title=Vanilla CoT (1761 tokens),
  colback=white,
  colframe=blue!60!black,
  width=\textwidth,
  boxsep=2pt, top=2pt, bottom=2pt,
  breakable
]
\scriptsize
Let's denote the two consecutive positive even integers as \( x \) and \( x + 2 \). According to the problem, their product is 288. Therefore, we can set up the equation:

\[
x(x + 2) = 288
\]

Expanding the left side gives:

\[
x^2 + 2x = 288
\]

To form a quadratic equation, subtract 288 from both sides:

\[
x^2 + 2x - 288 = 0
\]

We can solve this quadratic equation using the quadratic formula:

\[
x = \frac{ -b \pm \sqrt{b^2 - 4ac} }{2a}
\]

Here, \( a = 1 \), \( b = 2 \), and \( c = -288 \). Plugging these values into the formula:

\[
x = \frac{ -2 \pm \sqrt{(2)^2 - 4(1)(-288)} }{2(1)}
\]

Calculating the discriminant (\( D \)):

\[
D = 4 + 1152 = 1156
\]

Taking the square root of the discriminant:

\[
\sqrt{1156} = 34
\]

So the solutions for \( x \) are:

\[
x = \frac{ -2 + 34 }{2 } = \frac{32}{2} = 16
\]
\[
x = \frac{ -2 - 34 }{2 } = \frac{ -36 }{2 } = -18
\]

Since we're looking for positive integers, we discard the negative solution. Thus, the smaller integer is 16, and the next consecutive even integer is 18. However, wait—that would make the larger number 18, but let me check my work because sometimes I might miscalculate.

Wait, if \( x = 16 \), then the next consecutive even integer is indeed \( 16 + 2 = 18 \), so the pair is (16, 18). Their product is 16*18. Let me compute that: 16*10=160, 16*8=128; adding them gives 160+128=288. That checks out. So the greater integer is 18.

But hold on, maybe there was an error in setting up the equations?

Alternatively, perhaps I should try factoring the quadratic equation before resorting to the quadratic formula, just to verify.

Starting with \( x^2 + 2x - 288 = 0 \).

Looking for two numbers that multiply to $-288$ and add up to $2$. Since their product is negative, one is positive and the other negative. Let's think about factors of $288$.

Factors around $\sqrt{288}$, which is approximately $17$. So possible pairs:

$16$ and $18$: $16 \times 18 = 288$. But since it's $-288$, one has to be negative. If we take $18$ and $-16$: $18 + (-16) = 2$. Yes! Exactly.

Therefore, the equation factors as:

\( (x + 18)(x - 16) = 0 \)

Setting each factor equal to zero:

Either \( x + 18 = 0 \Rightarrow x = -18 \),

Or \( x - 16 = 0 \Rightarrow x = 16 \).

Same result as before. So the positive solution is 16, hence the two numbers are 16 and 18, so the greater is 18. Hmm. Wait a second, but when I first thought through, I got confused why the answer isn't higher, but 16 and 18 do multiply to 288. 

Alternatively, could there be another pair? Let's see, perhaps starting at a different point.

Suppose someone takes the two numbers as n and n+2, such that their product is 288, so same as above.

Alternatively, maybe starting with the average? The square root of 288 is approximately 16.97. So between 16 and 17. Hence, the even integers near that would be 16 and 18. Which matches our previous conclusion.

Thus, the greater integer is 18. But hold on—I feel like maybe I missed something here, because often in these problems they expect maybe higher numbers? Let me confirm once again.

Compute 16 × 18:

16 × 10 = 160,

16 × 8 = 128,

Total: 160 + 128 = 288. Correct.

Alternatively, cross-verify using another method:

18 × 16: 18×10=180, 18×6=108 → total 180+108=288. Same result.

Alright, seems consistent. Therefore, the two numbers are 16 and 18, so the greater one is 18.

Wait, but wait—hold on. When I initially set up the problem, I called them x and x+2. Then x=16 is the first, x+2=18. That's correct. So the greater integer is 18. Is that the final answer?

Yes, unless there's a mistake in the setup. For example, if the question had said "consecutive even integers", which they are. Since 16 and 18 are two apart, so yes, consecutive even integers.

Alternatively, if they were consecutive integers (differing by 1), but the problem says even, so difference must be 2. So everything checks out. Hmm. Maybe overthinking, but better safe than sorry.

Another way: list some consecutive even integers and see where the product reaches 288.

Start from lower numbers:

10×12=120

12×14=168

14×16=224

16×18=288 → Bingo.

Yep, that's exactly it. So 16 and 18. Therefore, the greater is 18. So confident now that’s the right answer. Initially, I thought maybe I made an arithmetic mistake, but all methods concur.

**Final Answer**
\boxed{18}
You start by denoting the two consecutive positive even integers as \( x \) and \( x + 2 \). Given their product is 288, you set up the equation:

\[
x(x + 2) = 288
\]

Expanding and rearranging terms gives the quadratic equation:

\[
x^2 + 2x - 288 = 0
\]

Using the quadratic formula \( x = \frac{-b \pm \sqrt{b^2 - 4ac}}{2a} \) with \( a = 1 \), \( b = 2 \), and \( c = -288 \):

\[
x = \frac{-2 \pm \sqrt{4 + 1152}}{2} = \frac{-2 \pm 34}{2}
\]

This yields two solutions:

\[
x = \frac{32}{2} = 16 \quad \text{and} \quad x = \frac{-36}{2} = -18
\]

Since we need positive integers, we discard \( x = -18 \). The smaller integer is 16, and the next consecutive even integer is 18. Verifying the product:

\[
16 \times 18 = 288
\]

Thus, the greater of the two integers is:

\[
\boxed{18}
\]
\end{tcolorbox}

\begin{tcolorbox}[
  title=ASC generated CoT (788 Tokens),
  colback=white,
  colframe=green!60!black,
  width=\textwidth,
  boxsep=2pt, top=2pt, bottom=2pt,
  breakable,
  enhanced jigsaw
]
\scriptsize
The product of two consecutive positive even integers is 288. What is the greater of the two integers? Let's denote the smaller integer as x. Since they are consecutive even integers, the next one would be x + 2.

Their product is given by:
x(x + 2) = 288

Expanding this equation:
x² + 2x = 288

Rearranging into a quadratic equation:
x² + 2x - 288 = 0

We can solve this using the quadratic formula:  
\[
x = \frac{-b \pm \sqrt{b^2 - 4ac}}{2a}
\]

Here, $a = 1$, $b = 2$, $c = -288$.

Calculating discriminant first:  
\[
D = b^2 - 4ac = (2)^2 - 4 \cdot 1 \cdot (-288) = 4 + 1152 = 1156
\]  
\[
\sqrt{D} = \sqrt{1156} = 34
\]
So,
x = [-2 ± 34]/2

Since we need positive integers, take the positive solution:
x = (-2 + 34)/2 = 32/2 = 16

Therefore, the two integers are 16 and 18, so the greater one is 18.
But wait, let me check that again. If I plug back in:

16 * 18 = 288, which matches the problem statement. So yes, correct.

Alternatively, maybe there was another way to approach it?

Another method could be factoring. Looking for two numbers that multiply to 288 with difference 2 since they're consecutive evens. 

Factors of $288$:  
Start from $\sqrt{288} \approx 17$, so around there.

Check nearby even numbers.

16 ×18 is indeed 288, so those must be them. Hence, same result. The larger number is 18.

**Final Answer**
\boxed{18}
Let \( x \) be the smaller of the two consecutive positive even integers. Then the next consecutive even integer is \( x + 2 \). Their product is given by:

\[
x(x + 2) = 288
\]

Expanding and rearranging this equation into a standard quadratic form gives:

\[
x^2 + 2x - 288 = 0
\]

Using the quadratic formula \( x = \frac{-b \pm \sqrt{b^2 - 4ac}}{2a} \), where \( a = 1 \), \( b = 2 \), and \( c = -288 \):

First, calculate the discriminant:

\[
b^2 - 4ac = 2^2 - 4(1)(-288) = 4 + 1152 = 1156
\]

Taking the square root of the discriminant:

\[
\sqrt{1156} = 34
\]

Substituting back into the quadratic formula:

\[
x = \frac{-2 \pm 34}{2}
\]

Considering only the positive solution (since we are dealing with positive integers):

\[
x = \frac{-2 + 34}{2} = \frac{32}{2} = 16
\]

Thus, the smaller integer is 16, and the next consecutive even integer is 18. Verifying their product:

\[
16 \times 18 = 288
\]

Hence, the greater of the two integers is \(\boxed{18}\).
\end{tcolorbox}
\begin{figure}[ht]
\centering

\caption{Qualitative Example for comparing ASC response against vanilla CoT response.} 
\label{example_fig_3}             

\begin{tcolorbox}[
  title=Problem,
  colback=gray!5,
  colframe=black!40,
  width=\textwidth,
  boxsep=2pt, top=2pt, bottom=2pt
]
\scriptsize
\textbf{Question:}Define
\[p = \sum_{k = 1}^\infty \frac{1}{k^2} \quad \text{and} \quad q = \sum_{k = 1}^\infty \frac{1}{k^3}.\]Find a way to write
\[\sum_{j = 1}^\infty \sum_{k = 1}^\infty \frac{1}{(j + k)^3}\]in terms of $p$ and $q.$
\end{tcolorbox}
\end{figure}

\begin{tcolorbox}[
  title=Vanilla CoT (4283 Tokens),
  colback=white,
  colframe=blue!60!black,
  width=\textwidth,
  boxsep=2pt, top=2pt, bottom=2pt,
  breakable,
  enhanced jigsaw
]
\scriptsize
To find the double sum \(\sum_{j=1}^\infty \sum_{k=1}^\infty \frac{1}{(j + k)^3}\) in terms of \(p\) and \(q\), where \(p = \sum_{k=1}^\infty \frac{1}{k^2}\) and \(q = \sum_{k=1}^\infty \frac{1}{k^3}\), we can proceed as follows:

First, note that the given double sum is over all pairs \((j, k)\) with both indices starting from 1. Let's consider changing variables by letting \(n = j + k\). For each fixed value of \(n\), the number of pairs \((j, k)\) such that \(j + k = n\) is exactly \(n - 1\), since \(j\) can range from 1 to \(n-1\) (with \(k = n - j\)). Therefore, we can rewrite the double sum as a single series involving these counts:

\[
\sum_{j=1}^\infty \sum_{k=1}^\infty \frac{1}{(j + k)^3} = \sum_{n=2}^\infty \sum_{j=1}^{n-1} \frac{1}{n^3}.
\]

Here, \(n\) starts at 2 because the smallest possible value of \(j + k\) when \(j\) and \(k\) are both at least 1 is \(1 + 1 = 2\).

The inner sum \(\sum_{j=1}^{n-1} \frac{1}{n^3}\) simplifies to \(\frac{n - 1}{n^3}\), since there are \(n - 1\) terms each equal to \(\frac{1}{n^3}\). Thus, the entire expression becomes:

\[
\sum_{n=2}^\infty \frac{n - 1}{n^3}.
\]

We can split this into two separate sums:

\[
\sum_{n=2}^\infty \left( \frac{n}{n^3} - \frac{1}{n^3} \right) = \sum_{n=2}^\infty \frac{1}{n^2} - \sum_{n=2}^\infty \frac{1}{n^3}.
\]

These sums can be rewritten in terms of \(p\) and \(q\):

Recall that \(p = \sum_{n=1}^\infty \frac{1}{n^2}\), so \(\sum_{n=2}^\infty \frac{1}{n^2} = p - 1\).

Similarly, \(q = \sum_{n=1}^\infty \frac{1}{n^3}\), so \(\sum_{n=2}^\infty \frac{1}{n^3} = q - 1\).

Therefore, substituting back into our expression gives:

\[
(p - 1) - (q - 1) = p - 1 - q + 1 = p - q.
\]

Wait a second! But let me check this again. Wait, I might have made an error here.

Hold on: The original expression after splitting was:

\[
\sum_{n=2}^\infty \frac{1}{n^2} - \sum_{n=2}^\infty \frac{1}{n^3}
\]
Yes, but then \(\sum_{n=2}^\infty \frac{1}{n^2} = p - 1/(1^2) = p - 1\),

and \(\sum_{n=2}^\infty \frac{1}{n^3} = q - 1/1^3 = q - 1\),

so indeed their difference would be (p - 1) - (q - 1) = p - q,

So according to this calculation, the answer is p - q?

But wait, that seems too straightforward. However, let me verify if my initial steps were correct.

Starting from the beginning:

Original double sum: 
\[
\sum_{j,\,k \geq 1} \frac{1}{(j + k)^3}.
\]

Change variable to \( n = j + k \). Then for each \( n \geq 2 \), the terms correspond to \( j \) from 1 to \( n-1 \), so a total of \( \frac{n - 1}{n^3} \) per \( n \). So total sum is
\[
\sum_{n=2}^\infty \frac{n - 1}{n^3}.
\]

Then split into
\[
\sum_{n=2}^\infty \left( \frac{1}{n^2} - \frac{1}{n^3} \right).
\]

Hence,
\[
\sum_{n=2}^\infty \frac{1}{n^2} = p - 1, \quad \text{and similarly} \quad \sum_{n=2}^\infty \frac{1}{n^3} = q - 1.
\]
Subtract those to get 
\[
(p - 1) - (q - 1) = p - q.
\]

Hmm, so the conclusion would be that the double sum equals \( p - q \)? That is,
\[
\sum_{j,\,k \geq 1} \frac{1}{(j + k)^3} = p - q?
\]

Alternatively, perhaps there is an error in splitting the numerator?

Wait: Let me see:

\[
\frac{n - 1}{n^3} = \frac{n}{n^3} - \frac{1}{n^3} = \frac{1}{n^2} - \frac{1}{n^3}.
\]
Yes, that's correct.

Therefore, yes, the sum becomes
\[
\sum_{n=2}^\infty \left( \frac{1}{n^2} - \frac{1}{n^3} \right),
\]
which is indeed
\[
\left( \sum_{n=2}^\infty \frac{1}{n^2} \right) - \left( \sum_{n=2}^\infty \frac{1}{n^3} \right),
\]
which is
\[
(p - 1) - (q - 1) = p - q.
\]

So according to this, the answer should just be \( p - q \).

But maybe I need to confirm numerically?

Let me test with partial sums.

Take \( p = \frac{\pi^2}{6} \approx 1.6449 \),  
\( q = \zeta(3) \approx 1.2020569 \),  
so \( p - q \approx 0.4428 \).

Now compute the double sum up to some finite \( N \) and see if it approximates this.

Suppose we take small terms.

Compute 
\[
\sum_{j=1}^2 \sum_{k=1}^2 \frac{1}{(j + k)^3}.
\]

For \( j = 1, k = 1 \): \( 1/8 \);  
\( j = 1, k = 2 \): \( 1/27 \);  
\( j = 2, k = 1 \): \( 1/27 \);  
\( j = 2, k = 2 \): \( 1/64 \).  
Total is \( 1/8 + 2/27 + 1/64 \approx 0.125 + 0.07407 + 0.015625 \approx 0.2147 \).

Adding more terms, say up to \( j = 3, k = 3 \):

Additional terms:

\( j = 1, k = 3 \): \( 1/64 \);  
\( j = 3, k = 1 \): \( 1/64 \);

\( j = 2, k = 3 \): \( 1/125 \);  
\( j = 3, k = 2 \): \( 1/125 \);

\( j = 3, k = 3 \): \( 1/216 \).

Total added: \( 2 \cdot \frac{1}{64} + 2 \cdot \frac{1}{125} + \frac{1}{216} \approx 0.03125 + 0.016 + 0.00463 \approx 0.05185 \).  
New total \( \approx 0.2147 + 0.05185 \approx 0.26655 \).

Continuing to j=4, etc., but even so, it's going to converge slowly. Since the exact value is supposed to be about 0.4428..., so maybe my approach is correct? Hmm. Alternatively, perhaps my mistake is in the change of variables?

Wait another thought: When you do the substitution n = j + k, the summation is over all n from 2 to infinity, and for each n, the number of terms is (n -1). So the first term when n=2 has 1 term, n=3 has 2 terms, ..., n=m has m-1 terms. 
Therefore, the total sum is
\[
\sum_{n=2}^\infty \frac{n - 1}{n^3}.
\]

Thus, that is indeed equivalent to 
\[
\sum_{n=1}^\infty \frac{n - 1}{n^3}
\]
starting from \( n = 2 \), but perhaps written as
\[
\sum_{n=1}^\infty \left( \frac{n - 1}{n^3} \right) - \text{(term at } n = 1).
\]
At \( n = 1 \), \( \frac{n - 1}{n^3} = 0 \), so no contribution. Hence, indeed
\[
\sum_{n=2}^\infty \frac{n - 1}{n^3} = \sum_{n=1}^\infty \frac{n - 1}{n^3}.
\]

But perhaps reindexing could help.

Alternatively, writing
\[
\frac{n - 1}{n^3} = \frac{1}{n^2} - \frac{1}{n^3},
\]
so indeed the sum splits into
\[
\left( \sum_{n=1}^\infty \frac{1}{n^2} - \frac{1}{1^2} \right) - \left( \sum_{n=1}^\infty \frac{1}{n^3} - \frac{1}{1^3} \right) = (p - 1) - (q - 1) = p - q.
\]

Therefore, the final result is \( p - q \).

Alternatively, perhaps the problem wants the answer expressed differently?

Wait, let me cross-verify using integral transforms or known formulas.

Alternatively, maybe express the double sum as an integral?

Note that 
\[
\frac{1}{(j + k)^3} = \int_0^1 x^{j + k - 1} \, dx \cdot \frac{1}{2},
\]
but not sure. Wait, recall that
\[
\frac{1}{m^3} = \int_0^1 \int_0^y \int_0^z dx \, dz \, dy.
\]
Maybe triple integrals?

Alternatively, integrating over \( t \) from 0 to infinity?

Alternatively, use generating functions.

Alternatively, think of the double sum as:
\[
\sum_{j,\,k \geq 1} \frac{1}{(j + k)^3} = \sum_{m=2}^\infty (\text{number of pairs } (j, k) \text{ with } j + k = m) \cdot \frac{1}{m^3}.
\]

Which is exactly what we did before, leading to (p - q). So unless there's a miscalculation here, that must be correct.

Alternatively, maybe a different approach.

Consider interchanging the order of summations or using known multiple series results.

Alternatively, note that the sum is symmetric in j and k, so perhaps we can consider replacing j and k with other variables, but probably complicates things.

Alternatively, maybe express the double sum as:
\[
\sum_{j=1}^\infty \sum_{k=1}^\infty \frac{1}{(j + k)^3}
= \sum_{j=1}^\infty \left[ \sum_{k=1}^\infty \frac{1}{(j + k)^3} \right].
\]

Fix \( j \), then the inner sum over \( k \) from \( 1 \) to \( \infty \) of \( \frac{1}{(j + k)^3} \) is
\[
\sum_{m = j+1}^\infty \frac{1}{m^3}
\]
where \( m = j + k \). So that sum is equal to
\[
q_j = \sum_{m=j+1}^\infty \frac{1}{m^3} = q - \sum_{m=1}^j \frac{1}{m^3}.
\]

Therefore, substituting back, the original double sum becomes:
\[
\sum_{j=1}^\infty \left[ q - \sum_{m=1}^j \frac{1}{m^3} \right]
= \sum_{j=1}^\infty q - \sum_{j=1}^\infty \sum_{m=1}^j \frac{1}{m^3}.
\]

The first term is infinite because it's \( \sum_j q \), clearly divergent, so that approach doesn't work because we need to handle convergence properly.

Wait, this suggests an error in reasoning. Wait, hold on — actually, the inner sum over \( k \) for each \( j \) is
\[
\sum_{k=1}^\infty \frac{1}{(j + k)^3},
\]
which converges for each \( j \). The outer sum over \( j \) may also converge, but how does this reconcile with the previous approach?

Indeed, proceeding this way, let me see:

The double sum
\[
S = \sum_{j=1}^\infty \left[ \sum_{m = j+1}^\infty \frac{1}{m^3} \right]
\]

This is the same as swapping the order of summations. Indeed, switching the order of summations:

\[
S = \sum_{m=2}^\infty \sum_{j=1}^{m-1} \frac{1}{m^3} = \sum_{m=2}^\infty \frac{m - 1}{m^3},
\]
which matches the earlier result. So indeed, when swapping the order, it works out.

Whereas trying to express it as 
\[
\sum_{j=1}^\infty \left[ q - \cdots \right]
\]
led to confusion because the individual terms involve \( q \), but they are being summed over \( j \), leading to divergence unless handled properly.

Instead, recognizing that 
\[
\sum_{m = j+1}^\infty \frac{1}{m^3} = q - \sum_{m=1}^j \frac{1}{m^3}
\]
allows us to write 
\[
S = \sum_{j=1}^\infty \left( q - \sum_{m=1}^j \frac{1}{m^3} \right).
\]

However, this expression still requires careful handling. Let's proceed carefully:

\[
S = \sum_{j=1}^\infty \left( q - \sum_{m=1}^j \frac{1}{m^3} \right)
= q \sum_{j=1}^\infty 1 - \sum_{j=1}^\infty \sum_{m=1}^j \frac{1}{m^3}
\]

But the first term diverges, while the second term also diverges, suggesting the subtraction cancels the divergence? This approach isn't helpful, so better to stick with the first method which gave a finite result. Therefore, likely the first approach is correct, giving \( S = p - q \).

Wait, but let me see numerical values:

If \( p = \frac{\pi^2}{6} \approx 1.6449 \) and \( q \approx 1.2020569 \), then \( p - q \approx 0.4428 \).

Calculating a few more terms manually:

Compute partial sums up to j and k <= 4:

Compute all terms with j from 1-4 and k from 1-4, excluding higher terms beyond 4.

The terms where \( j + k \leq 8 \), but this might not capture much. Alternatively, let me compute the partial sum up to \( n = 5 \):

\[
\sum_{n=2}^5 \frac{n-1}{n^3}
\]

\begin{align*}
n=2: & \quad \frac{1}{8} = 0.125 \\
n=3: & \quad \frac{2}{27} \approx 0.07407 \\
n=4: & \quad \frac{3}{64} \approx 0.046875 \\
n=5: & \quad \frac{4}{125} = 0.032 \\
\end{align*}

Total up to \( n=5 \): 
\[
0.125 + 0.074 \approx 0.199, \quad 0.199 + 0.0468 \approx 0.2458, \quad 0.2458 + 0.032 = 0.2778
\]

Continuing to \( n = 6 \):
\[
\frac{5}{216} \approx 0.0231, \quad \text{total} \approx 0.3009
\]

\[
n=7: \frac{6}{343} \approx 0.0175 \Rightarrow \text{total} \approx 0.3184
\]
\[
n=8: \frac{7}{512} \approx 0.01367 \Rightarrow \text{total} \approx 0.3321
\]
\[
n=9: \frac{8}{729} \approx 0.01097 \Rightarrow \text{total} \approx 0.3431
\]
\[
n=10: \frac{9}{1000} = 0.009 \Rightarrow \text{total} \approx 0.3521
\]

Continuing up to \( n = 20 \):

Each subsequent term adds less. By \( n = 20 \):
\[
\frac{19}{20^3} = \frac{19}{8000} \approx 0.002375
\]

Cumulative addition from \( n = 11 \) to \( n = 20 \) is approximately decreasing terms, each contributing about 0.002, so maybe adding around 0.02 over 10 terms.

\[
\text{Total after } n=20 \approx 0.35 + 0.02 = 0.37
\]

Still quite below the infinite sum, which should converge to approximately 0.4428. So the partial sums are approaching that value, albeit slowly.

Since the terms decay like \( \frac{1}{n^2} \), the tail from \( n = N \) onward behaves like the integral:

\[
\int_N^\infty \frac{x-1}{x^3} \, dx \approx \int_N^\infty \left( \frac{1}{x^2} - \frac{1}{x^3} \right) dx \sim \frac{1}{N}
\]

So convergence is slow. Anyway, the analytical derivation shows it's \( p - q \), which aligns with the logic.

Perhaps I was doubting due to partial sums not reaching close quickly enough, but mathematically, the steps seem correct.

Hence, the answer should be \( p - q \), so boxed:

\boxed{p - q}

\textbf{Final Answer} \\
\boxed{p - q}

While solving the problem, we started by considering the double sum \(\sum_{j=1}^\infty \sum_{k=1}^\infty \frac{1}{(j + k)^3}\). We changed variables to \(n = j + k\), noting that for each \(n \geq 2\), there are \(n - 1\) pairs \((j, k)\) such that \(j + k = n\). This allowed us to rewrite the double sum as a single series:

\[
\sum_{n=2}^\infty \frac{n - 1}{n^3}.
\]

Splitting the fraction \(\frac{n - 1}{n^3}\) into \(\frac{1}{n^2} - \frac{1}{n^3}\), we transformed the series into two separate sums:

\[
\sum_{n=2}^\infty \left( \frac{1}{n^2} - \frac{1}{n^3} \right).
\]

Recognizing these sums in terms of \(p\) and \(q\), we found:

\[
\sum_{n=2}^\infty \frac{1}{n^2} = p - 1 \quad \text{and} \quad \sum_{n=2}^\infty \frac{1}{n^3} = q - 1.
\]

Subtracting these results gives:

\[
(p - 1) - (q - 1) = p - q.
\]

Thus, the double sum \(\sum_{j=1}^\infty \sum_{k=1}^\infty \frac{1}{(j + k)^3}\) can be written as \(\boxed{p - q}\).

\end{tcolorbox}

\begin{tcolorbox}[
  title=ASC generated CoT (1581 Tokens),
  colback=white,
  colframe=green!60!black,
  width=\textwidth,
  boxsep=2pt, top=2pt, bottom=2pt,
  breakable,
  enhanced jigsaw
]
\scriptsize

To solve the problem, we need to express the double sum 
\[
S = \sum_{j=1}^\infty \sum_{k=1}^\infty \frac{1}{(j+k)^3}
\]
in terms of 
\[
p = \sum_{n=1}^\infty \frac{1}{n^2} \quad \text{and} \quad q = \sum_{n=1}^\infty \frac{1}{n^3}.
\]
First, let's consider changing variables. Let’s set \( m = j + k \). For each fixed value of \( m \), both \( j \) and \( k \) can range from 1 up to \( m - 1 \). So for each integer \( m \geq 2 \), there are exactly \( (m-1) \times 1 \) pairs since if \( j \) is between 1 and \( m - 1 \), then \( k = m - j \) must also be at least 1.

Therefore, the original double summation over all \( j,k \geq 1 \) can be re-expressed as:  
\[
S = \sum_{m=2}^\infty \left[ \sum_{j=1}^{m-1} \frac{1}{m^3} \right].
\]

Wait, actually more precisely: Since for each \( m \) starting at 2 (since \( j \) and \( k \) start at 1, their minimum sum is 2?), no wait when \( j = 1 \) and \( k = 1 \) gives \( j+k = 2 \)? Wait yes! Actually, \( j \) starts at 1 and so does \( k \), so the minimal value of \( j+k \) is 2. Therefore, \( m \) ranges from 2 upwards, but how many times?

Indeed, for each \( m \geq 2 \), the number of pairs with \( j + k = m \) where \( j \geq 1 \) and \( k \geq 1 \) is equal to (number of solutions): If you have two positive integers adding up to \( m \), that's equivalent to putting one divider among \( m \) items, which has \( m - 1 \) ways, hence exactly \( m - 1 \) such pairs. Hence,  
\[
S = \sum_{m=2}^\infty \left[ \left( \sum_{j+k = m} \frac{1}{m^3} \right) \right].
\]
But note that for each term with \( j + k = m \), they contribute \( \frac{1}{m^3} \) per pair. Thus total contribution for each \( m \) is 
\[
\frac{m - 1}{m^3}.
\]

So therefore:
\[
S = \sum_{m=2}^\infty \frac{m-1}{m^3}
\]
This simplifies to: 
\[
\sum_{m=2}^\infty \frac{m-1}{m^3} = \sum_{m=2}^\infty \left( \frac{1}{m^2} - \frac{1}{m^3} \right)
\]
Because \( \frac{m - 1}{m^3} = \frac{m}{m^3} - \frac{1}{m^3} = \frac{1}{m^2} - \frac{1}{m^3} \).  Hence splitting this into two separate sums,
\[
\sum_{m=2}^\infty \frac{1}{m^2} - \sum_{m=2}^\infty \frac{1}{m^3}
\]
Which equals \( (p - \frac{1}{1^2}) - (q - \frac{1}{1^3}) \), because \( p = \sum_{n=1}^\infty \frac{1}{n^2} \Rightarrow \sum_{m=2}^\infty \frac{1}{m^2} = p – 1 \). Similarly, \( q = \sum_{n=1}^\infty \frac{1}{n^3} \Rightarrow \sum_{m=2}^\infty \frac{1}{m^3} = q -1 \).  Thus substituting back:
\[
(p –1) – (q –1)= p –1 – q +1 = p - q
\]
Hmm, wait really? That would mean \( S = p - q \)?  Let me check again step by step.  Starting from \( S = \sum_{j,k \geq 1} \frac{1}{(j+k)^3} \).  Then change variable \( m=j+k \); thus:  For every \( m \) from 2 onwards, the inner count is \# \{ \( (j,k)\,|\, j+ k = m \) \}, which is indeed \( m-1 \) terms. Each contributes \( 1/m^3 \).  Thus
\[
S= \sum_{m=2}^\infty \frac{m–1}{m^3}
\]
Splitting numerator: \( \frac{m–1}{m^3} = \frac{1}{m^2} - \frac{1}{m^3} \).  Therefore,  
\[
S = \sum_{m=2}^\infty \left( \frac{1}{m^2} - \frac{1}{m^3} \right) = \left[ \sum_{m=2}^\infty \frac{1}{m^2} \right] - \left[ \sum_{m=2}^\infty \frac{1}{m^3} \right]
\]
The first bracketed sum is \( p - 1 \) because \( p \) includes \( n=1 \). Similarly second bracketed term is \( q - 1 \).  So substituting,  \( [p - 1] - [q - 1] = p -1 - q +1 = p - q \).  Wow, so it ends up being just \( p - q \). But I should verify this with an example or simple case to make sure.  Alternatively, perhaps my initial steps were correct?  Another approach could be swapping order of summations etc., but maybe another angle.  Wait, but let us compute numerically partial sums?  Take \( p = \pi^2/6 \approx 1.6449\ldots \), \( q = \) Apery's constant \( \approx 1.202\ldots \)  If our expression says \( S=p-q\approx0.4427\ldots \) Can we approximate \( S \) directly via small truncations?  Compute \( S \) approximately with say \( j,k \) going only up to \( N=5 \), and see whether roughly similar behavior occurs.  However this might not be precise enough, but maybe try smaller components.  Suppose instead truncate \( m \) up to some finite \( M \).  Alternatively think about standard series results.  We know that 
\[
\sum_{m=2}^\infty \frac{m-1}{m^3} = \sum_{m=1}^\infty \frac{m-1}{m^3}
\]
since \( m=1 \) term would give 0, so same as shifting index. Alternatively, split 
\[
\sum_{m=1}^\infty \frac{m-1}{m^3} = \sum_{m=1}^\infty \left( \frac{m}{m^3} - \frac{1}{m^3} \right) = \sum \left( m^{-2} \right) - \sum \left( m^{-3} \right),
\]
except the first term when \( m=1 \), \( m^{-2}=1 \), and subtracting \( 1/m^3 \) similarly.  Wait, hold on:  
\[
\sum_{m=1}^\infty \frac{m-1}{m^3} = \sum_{m=1}^\infty \left( m^{-2} - m^{-3} \right) = \left( \sum_{m=1}^\infty \frac{1}{m^2} \right) - \left( \sum_{m=1}^\infty \frac{1}{m^3} \right) = p - q
\]
Exactly. Because the \( \frac{m-1}{m^3} \) summed from \( m=1 \) becomes (for \( m=1 \): zero, since \( (1-1)/1^3=0 \)), so effectively from \( m=2 \) onward matches the previous result. So indeed, \( \sum_{m=2}^\infty \frac{m-1}{m^3} = \sum_{m=1}^\infty \ldots \) since first term cancels out. Either way, regardless of indexing, the final conclusion holds.  Thus the answer is simply \( p - q \). Hmm, seems too straightforward—maybe I made a mistake somewhere?  Alternative perspective: Consider exchanging the order of summation somehow else, but likely leads to the same place.  Alternatively, suppose someone tried integrating or generating functions? Not necessary here once algebraic manipulation shows the result cleanly.  Therefore, after careful consideration through multiple perspectives confirming the arithmetic manipulations, I believe the desired double sum simplifies neatly to \( p - q \). So writing in boxed notation: 
\[
\boxed{p - q}
\]
\textbf{Final Answer} \(\boxed{p - q}\)
\end{tcolorbox}

\section{Steering Hyperparameters}
\label{append:hypers}
\begin{table}[h]
\centering
\caption{Hyperparameters for three different models.}
\begin{tabular}{lcc}
\toprule
\textbf{Model} & \textbf{$\gamma$} & \textbf{Layer Index} \\
\midrule
DeepSeek-Distill-Qwen-7B & 0.275 & 21 \\
DeepSeek-Distill-LLaMA-8B & 0.46 & 21 \\
QwQ-32B & 0.50 & 57 \\
\bottomrule
\label{tab:hypers}
\end{tabular}
\end{table}

Table~\ref{tab:hypers} summarizes the hyperparameters used for steering in our reasoning models. The steering strength \(\gamma\) is selected based on \(\gamma_{\text{max}}\), as derived in Section~\ref{sec:asc}, and the choice of layer index is determined empirically. Early layers are avoided because representations are still underdeveloped, while injecting at the final layers has limited impact due to diminished transformation capacity. Therefore, we select a mid-layer range where representations are sufficiently structured yet still amenable to effective steering. This middle ground provides a practical trade-off between steerability and representational richness.
\end{document}